\documentclass{article} 
\usepackage{iclr2025_conference,times}

\usepackage{amsmath,amsfonts,bm}

\def\eqref#1{equation~\ref{#1}}

\def\1{\bm{1}}

\DeclareMathAlphabet{\mathsfit}{\encodingdefault}{\sfdefault}{m}{sl}
\SetMathAlphabet{\mathsfit}{bold}{\encodingdefault}{\sfdefault}{bx}{n}

\DeclareMathOperator*{\argmin}{arg\,min}

\usepackage{hyperref}
\usepackage{url}
\usepackage{graphicx}
\usepackage{subcaption}
\usepackage{todonotes}

\title{Low Stein Discrepancy via Message-Passing Monte Carlo}

\author{Nathan Kirk \\ 
Illinois Institute of Technology\\
\And T. Konstantin Rusch\\
MIT
\And Jakob Zech \\
Heidelberg University
\And
Daniela Rus \\
MIT\\
}

\iclrfinalcopy 
\begin{document}

\maketitle

\begin{abstract}
Message-Passing Monte Carlo (MPMC) was recently introduced as a novel low-discrepancy sampling approach leveraging tools from geometric deep learning. While originally designed for generating uniform point sets, we extend this framework to sample from general multivariate probability distributions $F$ with known probability density function. Our proposed method, Stein-Message-Passing Monte Carlo (Stein-MPMC), minimizes a kernelized Stein discrepancy, ensuring improved sample quality. Finally, we show that Stein-MPMC outperforms competing methods, such as Stein Variational Gradient Descent and (greedy) Stein Points, by achieving a lower Stein discrepancy.
\end{abstract}

\section{Introduction}
Approximating a probability distribution with a discrete set of points is 
a fundamental task
in modern scientific computation with wide ranging applications, examples of which include uncertainty quantification, Bayesian inference, and numerical integration. 
All of these problems correspond to computing expectations of the form $\mathbb{E}_f(q)$ of a function $q(\boldsymbol{x})$ in $\mathbb{R}^d$ with respect to a given distribution $F$ with probability density function $f(\boldsymbol{x})$. Monte Carlo (MC) methods are a popular choice for approximating the integral by the sample mean of $q$ evaluated on a set of $N$ sample nodes $\{\mathbf{X}_i\}_{i=1}^N$ drawn IID from distribution $F$, i.e., 

\begin{equation}\label{eq:problem}
    \mathbb{E}_f(q) = \int_{{\mathbb{R}^d}} q(\boldsymbol{x}) f(\boldsymbol{x}) \mathrm{d}\boldsymbol{x} = \int_{\mathbb{R}^d} q(\boldsymbol{x}) \mathrm{d} F(\boldsymbol{x}) \approx \frac{1}{N} \sum_{i=1}^N q(\mathbf{X}_i)
\end{equation}
for $\mathbf{X}_i \overset{\text{IID}}{\sim} F$. Provided that the variance of the integrand is bounded, the standard Monte Carlo rate of $\mathcal{O}(N^{-1/2})$ applies often necessitating a very large $N$ when high precision is required. Therefore, to obtain greater accuracy, or a faster convergence rate, one may replace the random evaluations by a carefully chosen deterministic set that better represents the distribution $F$. These so-called \textit{low-discrepancy} points form the basis of quadrature rules that fall under the umbrella of quasi-Monte Carlo (QMC) methods \cite{QMCsurvey13, QMCsurvey25}.

Assume that the function $q$ belongs to a reproducing kernel Hilbert space (RKHS) $\mathcal{H}$ of functions from \( \mathbb{R}^d \rightarrow \mathbb{R} \) equipped with an inner product \(\langle \cdot, \cdot \rangle_{\mathcal{H}}\) and corresponding norm \(\| \cdot \|_{\mathcal{H}}\). One can then use the Cauchy-Schwarz inequality within \(\mathcal{H}\) to derive an error bound on the approximation (\ref{eq:problem}) as
\[
\left| \frac{1}{N} \sum_{i=1}^N q(\mathbf{X}_i) - \int q \, dF \right| \leq \| q \|_{\mathcal{H}} \, D_{\mathcal{H}, F} \left( \{\mathbf{X}_i\}_{i=1}^N \right).
\]
In the above, \( D_{\mathcal{H}, F} \left( \{\mathbf{X}_i\}_{i=1}^N \right) \) is referred to as the \textit{discrepancy}, and the term $\| q \|_{\mathcal{H}}$ is a measure of variation of the integrand; see \cite{hickernell1998generalized} for further details. The discrepancy term measures how closely the empirical distribution of the discrete sample point set approximates the target distribution $F$. Denote by $k:\mathbb{R}^d\times\mathbb{R}^d\to\mathbb{R}$ the reproducing kernel associated with the RKHS $\mathcal{H}$. When both the integral \( k_F := \int k(x, \cdot) \, dF(x) \in \mathcal{H} \) and \( k_{F, F} := \int k_F \, dF \) are explicitly available, the discrepancy can be calculated directly by
\begin{equation}\label{eq:general_disc}
D_{\mathcal{H}, F} \left( \{\mathbf{X}_i\}_{i=1}^N \right) = \sqrt{k_{F, F} - \frac{2}{N} \sum_{i=1}^N k_F(\mathbf{X}_i) + \frac{1}{N^2} \sum_{i,j=1}^N k(\mathbf{X}_i, \mathbf{X}_j)}.
\end{equation}

The case when $F$ is the uniform distribution is very well studied and several classical and computable measures of discrepancy exist, along with many known constructions of uniform low-discrepancy point sets and sequences; see \cite{KUIPNIED1974, Dick_Pillichshammer_2010, LATTICESBOOK}. For a nonuniform distribution $F$, computable discrepancy measures and corresponding low-discrepancy point sets are not as widespread. Thus, hoping to exploit the existing constructions for $U[0,1]^d$, there exist several transformations to map uniform low-discrepancy points to a nonuniform distribution $F$. For Gaussians, the Box-Muller transformation \cite{BOXMULLER1958} provides a direct and easily computable transport map coupling the uniform distribution with the target $F$. For general distributions $F$, in one dimension, the inverse CDF provides such a transport, while the Rosenblatt transformation \cite{ROSENBLATT1952} extends this approach to higher dimensions. Numerous methods have been developed to compute transport maps in practice, including normalizing flows \cite{pmlr-v37-rezende15}, neural ODEs \cite{NEURIPS2018_69386f6b}, or polynomial transports \cite{marzouk2016introduction}. Each of these approaches comes with its own challenges, mostly related to solving highly nonconvex optimization problems. Overall, it is desirable to be able to effectively generate low-discrepancy samples directly from a target distribution $F$.

When $F$ is not the uniform distribution, direct optimization of the discrepancy (\eqref{eq:general_disc}) can be a difficult problem without a clear, efficient objective function to minimize. In recent years, there has emerged an active area of research on this topic using variants of a procedure derived from Stein's method \cite{STEIN1972}; see \cite{GORHAMMACKEY2017, StochasticSD2020, MINSteinDisc2019, STEINPOINTS, SVGDwithoutgrad2018, SVGD2016, LIULEEJORDAN2016, SVGDasgradientflow2017, GradientFreeCSD2023, GradientFreeKSD2024} and references therein. The papers \cite{GORHAMMACKEY2017, CHWGRETTON2016, LIULEEJORDAN2016} independently introduced the \emph{kernel Stein discrepancy (KSD)}, one of several computable versions of the Stein discrepancy, which is used to assess the ``closeness" of a sample point set to a target distribution $F$. 


\subsection{Our Contribution}

In this paper, we extend the Message-Passing Monte Carlo \cite{RUSCHKIRK2024} framework to minimize a kernelized Stein discrepancy, ensuring improved sample quality from general multivariate probability distributions $F$ with known probability density function. We compare Stein Discrepancy values against the point sets generated by the benchmark methods of Stein Variational Gradient Descent \cite{SVGD2016} and Stein Points \cite{STEINPOINTS}.

\section{Stein Discrepancy}

The work of \cite{GorhamMackey2015} introduced a new family of sample quality measures, known as the Stein discrepancies, which can be used to measure the error in the approximation (\ref{eq:problem}) without explicitly integrating under $F$. Stein discrepancies are derived using Stein’s identity \cite{STEIN1972}, a key result in probability theory that relates the expectation of a derivative-based function to properties of a target distribution. There exist several computable versions of the Stein discrepancy family, e.g., graph Stein discrepancies. However, kernel Stein discrepancies have gained most attention due to their closed-form expression involving the sum of kernel evaluations over pairs of sample points.

More formally, for a Stein operator \(T_F\), the following holds
\[
\int T_F[p](x) \, dF(x) = 0 \quad \forall p \in \mathcal{F},
\]
where \(\mathcal{F}\) is a set of functions that are sufficiently smooth. Stein’s identity allows the construction of such operators \(T_F\) that characterize how well a distribution matches a target. When \(\mathcal{F}\) is chosen as a RKHS \(\mathcal{H}\) with a reproducing kernel \(k\), the image of \(\mathcal{H}\) under \(T_F\) is denoted as \(\mathcal{H}_0 = T_F \mathcal{H}\). The KSD is then computed using the reproducing kernel \(k_0\) of \(\mathcal{H}_0\), defined as
\[
k_0(x, x') = T_F T_F^* k(x, x'),
\]
where \(T_F^*\) is the adjoint of the Stein operator \(T_F\), acting on the second argument of the kernel. A common choice for \(T_F\) is the Langevin Stein operator
defined by
\[
T_F p(x) = \nabla \cdot (f(x) p(x)) / f(x),
\]
where \(\nabla \cdot\) is the divergence operator, and \(p\) is a vector-valued function in the RKHS \(\mathcal{H}^d\). This operator leads to the Stein reproducing kernel
\[
\begin{aligned}
k_0(x, x') = & \nabla_x \cdot \nabla_{x'} k(x, x') + \nabla_x k(x, x') \cdot \nabla_{x'} \log f(x') \\ &+ \nabla_{x'} k(x, x') \cdot \nabla_x \log f(x) + k(x, x') \nabla_x \log f(x) \cdot \nabla_{x'} \log f(x').
\end{aligned}\label{eq:Stein_repro_kernel}
\]
Stein kernels possess the nice property that $k_{0,F} = \int k_0(x,\cdot)dF = 0$ and $k_{0,F,F} = \int k_{0,F}dF = 0$. Thus for some base kernel $k$, the KSD is computed from \eqref{eq:general_disc} as
\begin{equation}\label{eq:KSD}
D_{\mathcal{H}_0, F}(\{\mathbf{X}_i\}_{i=1}^N) = \sqrt{\frac{1}{N^2} \sum_{i, j=1}^N k_0(\mathbf{X}_i, \mathbf{X}_j)}.
\end{equation}


\section{Stein-Message-Passing Monte Carlo (Stein-MPMC)}

\begin{figure}
    \centering
    \includegraphics[width=0.8\textwidth]{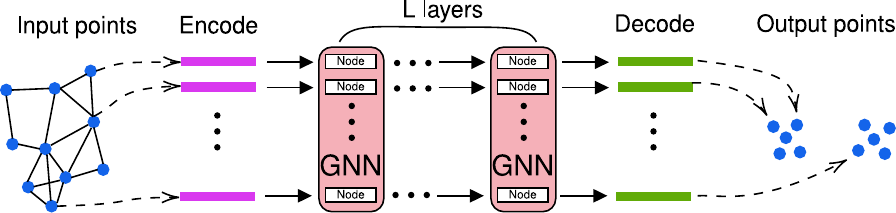}
    \caption{Schematic of the MPMC model. The points are encoded to a high dimensional representation then passed through multiple layers of a message-passing GNN where the underlying computational graph is constructed based on nearest neighbors. Finally, the node-wise output representations of the final layer of the GNN are decoded.}
    \label{fig:MPMCmodel}
\end{figure}

Message-Passing Monte Carlo (MPMC) \cite{RUSCHKIRK2024} represents a significant advancement in the field of quasi-Monte Carlo methods and general low-discrepancy sampling applications \cite{chahine2024improving}. MPMC leverages tools from geometric deep learning, including graph neural networks (GNNs) and a message-passing framework, to effectively learn a transformation mapping random input point set to uniform low-discrepancy points in the $d$-dimensional unit hypercube. In the original MPMC framework (see Figure \ref{fig:MPMCmodel}), the target is always the uniform distribution on the $d-$dimensional hypercube and the training is guided by Warnock's formula \cite{warnock1972computational} for the $L_2$-discrepancy -- a classical measure of uniformity for sample point sets in $[0,1]^d$. In the proposed Stein-MPMC model, as described below, the architecture is holistically similar, with the primary change being the Stein discrepancy based objective function.


\subsection{Stein-MPMC Model}

Our objective is to train a neural network to effectively learn a mapping to transform an initialized sample point set $\{\mathbf{X}\}_{i=1}^N$ into points $\{\hat{\mathbf{X}}\}_{i=1}^N$ that minimize the (kernel) Stein discrepancy (\ref{eq:KSD}) where $\mathbf{X}_i,\hat{\mathbf{X}}_i \in \mathbb{R}^d$ for all $1\leq i \leq N$. The input point set will be generated randomly from target distribution $F$ where possible, i.e., $\mathbf{X}_i \overset{\text{IID}}{\sim} F$ for $1\leq i \leq N$. However, in principle, our initialized training data can be taken as any reasonable set of points not judiciously chosen to be purposefully far from the target $F$.

For the model architecture, we start by constructing an undirected computational graph $G = (V, E \subseteq V \times V)$, where $V$ denotes the set of unordered nodes corresponding to the input points $\{\mathbf{X}_i\}_{i=1}^N$, and $E$ is the set of pair-wise connections between the nodes. We denote the $1$-neighborhood of a node $i \in V$ as 
$\mathcal{N}_i = \{j \in V : (i,j) \in E\}$ and set $$\mathcal{N}_i = \{j \in V : \| \mathbf{X}_i -\mathbf{X}_j \|_2 \leq r\}$$ for a fixed radius $r \in \mathbb{R}$. That is, every node $i \in V$ is connected to every other node $j \in V$ that is within a neighborhood of radius $r$ of node $i$. This local connectivity of nodes emphasizes that the network should primarily consider near-by points when learning the transformation. The main aspect of the Stein-MPMC model is the GNN layers based on the message-passing framework. Message-passing GNNs are a family of parametric functions defined through local updates of hidden node representations. More concretely, we iteratively update node features as,
\[
\mathbf{X}_i^{l} = \phi^l \left( \mathbf{X}_i^{l-1}, \sum_{j \in \mathcal{N}_i} \psi^l (\mathbf{X}_i^{l-1}, \mathbf{X}_j^{l-1}) \right), \quad \text{for all } l = 1, \dots, L,
\]
with \( \mathbf{X}_i^{l} \in \mathbb{R}^{m_l} \) for all nodes \( i \). Moreover, we parameterize \( \phi^l, \psi^l \) as ReLU-multilayer perceptrons (MLPs), i.e., MLPs using the element-wise ReLU activation function, $\text{ReLU}(x) = \max(0, x)$, in-between layers. We further encode the initial node features by an affine transformation, $\mathbf{X}_i^{0} = \mathbf{A}_{\text{enc}} \mathbf{X}_i + \mathbf{b}_{\text{enc}}$ for all \( i = 1, \dots, N \), with weight matrix \( \mathbf{A}_{\text{enc}} \in \mathbb{R}^{m_0 \times d} \) and bias \( \mathbf{b}_{\text{enc}} \in \mathbb{R}^{m_0} \).  Finally, we decode the output of the final GNN layer by an affine transformation back into \(\mathbb{R}^d\), i.e., $
\hat{\mathbf{X}}_i = \mathbf{A}_{\text{dec}} \mathbf{X}_i^L + \mathbf{b}_{\text{dec}}$ for all \( i = 1, \dots, N \), with the weight matrix \( \mathbf{A}_{\text{dec}} \in \mathbb{R}^{d \times m_L} \), and bias \( \mathbf{b}_{\text{dec}} \in \mathbb{R}^{d} \).

Lastly, the training objective is selected to be the kernel Stein discrepancy (\ref{eq:KSD}). Reasons for this choice are two-fold; i) the kernelized version of Stein discrepancy has closed form and fast parallelizable computation of kernel evaluations of pairs of points, and ii) for carefully chosen base kernels $k$ in \eqref{eq:KSD}, there exist results that the KSD controls weak convergence to the target distribution $F$; see \cite{STEINPOINTS} and \cite[Theorem 8]{GORHAMMACKEY2017}.

\section{Results}

The proposed Stein-MPMC method is empirically assessed and compared against existing benchmark methods. Precisely, we illustrate across two examples of target distribution $F$ that Stein-MPMC generates sample point sets with a smaller KSD with respect to $F$. In two dimensions only, we examine a Gaussian mixture over the unbounded domain $\mathbb{R}^2$, and a distribution defined as the product of two independent Beta distributions over the unit square.

\subsection{Experimental Detail}

For the base kernel function $k$ in \eqref{eq:Stein_repro_kernel}, we use the standard Radial Basis Function (RBF) kernel
\[
k(x,x') = \exp \left( - \frac{\|x-x'\|^2}{2h^2} \right).
\]
where the bandwidth parameter $h$ is not tuned separately for different experiments or methods and instead, we apply the median heuristic across all experiments. Specifically, we take we take the bandwidth to be $h = \sqrt{\text{med}^2/2\log (N+1)}$ where $\text{med}$ is the median of the pairwise distances between the current sample point set. This choice is motivated to ensure a fair comparison across methods and prevent confounding effects due to kernel tuning.

We compare Stein-MPMC against two established methods:

\begin{enumerate}
    \item Stein Variational Gradient Descent (SVGD) \cite{LIULEEJORDAN2016} generates point sample sets from a target distribution $F$ by performing a version of gradient descent on the Kullback-Leibler (KL) divergence $\text{KL}( \cdot \| F )$. Like Stein-MPMC, SVGD is a global optimization method that updates all sample points simultaneously at each step.
    \item Stein points \cite{STEINPOINTS} generates sequences of points from a distribution $F$ via greedy minimization of the KSD directly. Despite being a greedy sequential procedure, rather than the global methods of SVGD and Stein-MPMC, Stein points is included in the comparison due to its direct KSD optimization.
\end{enumerate}

We generate samples using Stein-MPMC, SVGD, and Stein Points for values of $N$ ranging from $20$ to $500$ in increments of $40$, recording the corresponding KSD values. For the greedy Stein Points method, we track the KSD value at each instance of $N$ during a single sequential run as the sample set grows to $500$ total points.

Full experimental details including optimization methods and lists of selected hyperparameters for each method are given in Appendix \ref{app:experimentaldetail}.

\begin{figure}[t]
    \centering
    \begin{subfigure}[b]{0.48\textwidth} 
        \centering
        \includegraphics[width=\textwidth]{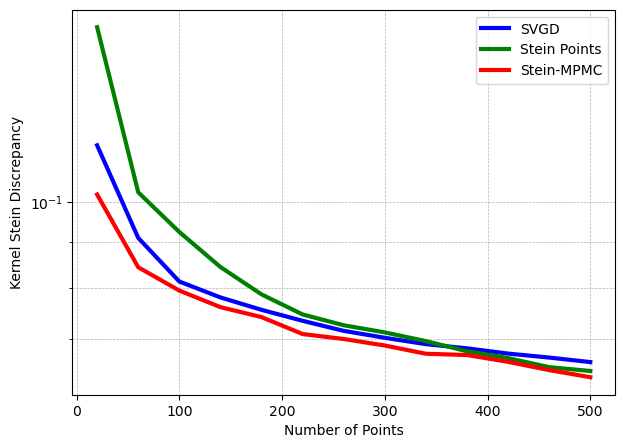}
        \caption{Gaussian Mixture}
        \label{fig:ksd_gaussian}
    \end{subfigure}
    \hfill
    \begin{subfigure}[b]{0.48\textwidth}  
        \centering
        \includegraphics[width=\textwidth]{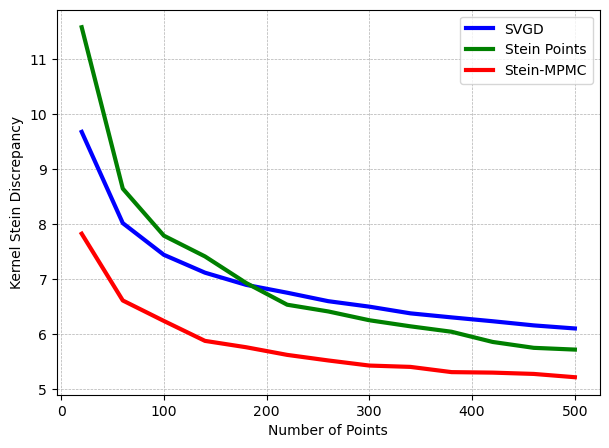}
        \caption{Beta Product}
        \label{fig:beta_results}
    \end{subfigure}
    \caption{KSD results for our two target distributions. Stein-MPMC yields smaller KSD values for every $N = 20, 60, 100, \ldots, 500$ across both distributions.}
    \label{fig:results}
\end{figure}

\subsection{Gaussian Mixture Distribution}
We first consider a Gaussian mixture model in two dimensions, which is somewhat of a standard benchmark for variational inference methods. The target distribution is a two-component Gaussian mixture
\[
\frac{1}{2} \mathcal{N}(\mu_1, \Sigma_1) + \frac{1}{2} \mathcal{N}(\mu_2, \Sigma_2),
\]
where $\mu_1 = (-1.5,0), \mu_2 = (1.5,0), \Sigma_1 = \Sigma_2 = I$.
KSD values for the three methods are shown in Figure \ref{fig:ksd_gaussian}, demonstrating that Stein-MPMC outperforms Stein Points and SVGD. For this Gaussian mixture example, as $N$ increases the differences between the methods become less pronounced.

\subsection{Beta Product Distribution}

We also consider a Beta-distributed target density as an example of a bounded probability distribution. The target distribution is defined as the product of two independent Beta distributions
\[
X \sim \text{Beta}(\alpha_x, \beta_x), \quad Y \sim \text{Beta}(\alpha_y, \beta_y).
\]
This distribution is supported on \( (0,1) \times (0,1) \) and allows independent control over the shape of each marginal through the parameters \( \alpha_x, \beta_x, \alpha_y, \beta_y \). For our experiments, we set $\alpha_x = 2, \beta_x = 4, \alpha_y = 2, \beta_y = 4$. Discrepancy values for each method for this Beta distribution are given in Figure \ref{fig:beta_results} and are consistent with those of the Gaussian mixture example; Stein-MPMC outperforms the other methods with respect to KSD values for all tested instances of $N$.

\section{Discussion}

Stein-MPMC effectively minimizes kernel Stein discrepancy, outperforming SVGD and Stein Points by leveraging message-passing graph neural networks to solve the nonconvex global optimization problem. The results support the existing notion that optimizing sample point distributions for a pre-determined $N$ allows for better sample uniformity compared to sequential generation. Future work should explore its scalability to higher dimensions and the impact of adaptive kernel tuning to further enhance sample quality.


\subsubsection*{Acknowledgments}
The work of NK was supported by the National Science Foundation (DMS Grant No. 2316011). NK also gratefully acknowledges the Argonne Leadership Computing Facility (ALCF) for providing GPU access which supported the computational aspects of this research. This work was supported in part by the Postdoc.Mobility grant P500PT-217915 from the Swiss National Science Foundation, the Schmidt AI2050 program (grant G-22-63172), and the Department of the Air Force Artificial Intelligence Accelerator and was accomplished under Cooperative Agreement Number FA8750-19-2-1000. The views and conclusions contained in this document are those of the authors and should not be interpreted as representing the official policies, either expressed or implied, of the Department of the Air Force or the U.S. Government. The U.S. Government is authorized to reproduce and distribute reprints for Government purposes notwithstanding any copyright notation herein.

\bibliography{refs}
\bibliographystyle{iclr2025_conference}


\newpage
\appendix

\section{Training Details}\label{app:experimentaldetail}

All experiments have been run on NVIDIA DGX A100 GPUs.

\subsection{Stein-MPMC}

 Each model was trained for 50k epochs with the Adam optimizer \cite{adamopt}. Stein-MPMC hyperparameters were tuned using \texttt{Optuna} Python package \cite{optuna} random search over the search spaces and distributions contained in Table \ref{tab:hyperparameter_search}.

\begin{table}[h]
    \centering
    \renewcommand{\arraystretch}{1.2}
    \begin{tabular}{l c c}
        \hline
        \textbf{Hyperparameter} & \textbf{Range} & \textbf{Distribution} \\
        \hline
        learning rate & $[10^{-4}, 10^{-2}]$ & log uniform \\
        hidden size $m_0 = m_1 = \dots = m_L$ & $\{32, 64, 128, 256\}$ & uniform \\
        number of GNN layers $L$ & $\{1, 2, 3,4, 5\}$ & uniform \\
        weight decay & $[10^{-6}, 10^{-2}]$ & log uniform \\
        \hline
    \end{tabular}
    \caption{Hyperparameter search-space and respective random distributions.}
    \label{tab:hyperparameter_search}
\end{table}

\subsection{Stein Points}

Computation of the $N^{th}$ Stein point is dependent upon the already existing $N-1$ terms and requires a global optimization to find $\mathbf{X}_N \in \mathbb{R}^d$ that minimizes the kernel Stein discrepancy of the total $N$ element set, holding $\{\mathbf{X}_i\}_{i=1}^{N-1}$ fixed. In \cite{STEINPOINTS}, several numerical optimization methods are considered to solve this $\argmin$ problem. In our experiments, we implement the Adam optimizer with a learning rate of $0.01$, selected after testing several judiciously chosen alternatives for the learning rate.

\subsection{Stein Variational Gradient Descent}

The other global optimization method considered was Stein Variational Gradient Descent introduced in \cite{SVGD2016}. SVGD was trained with the standard update rule 
\[
\mathbf{X}_i^{(t+1)} = \mathbf{X}_i^{(t)} + \eta \left( \frac{1}{n} \sum_{j=1}^{n} k(\mathbf{X}_i, \mathbf{X}_j) \nabla \log f(\mathbf{X}_j) + \nabla_{\mathbf{X}_j} k(\mathbf{X}_i, \mathbf{X}_j) \right)
\]
where $k$ is the base kernel (taken to be the RBF kernel with median bandwidth), step size $\eta$ was fixed at $0.001$ and was run for a standard $50k$ iterations on each experiment.

\end{document}